# Dual Embodied-Symbolic Concept Representations for Deep Learning

Daniel T. Chang (张遵)

*IBM (Retired)* dtchang43@gmail.com

**Abstract:** Motivated by recent findings from cognitive neural science, we advocate the use of a dual-level model for concept representations: the embodied level consists of concept-oriented feature representations, and the symbolic level consists of concept graphs. Embodied concept representations are modality specific and exist in the form of feature vectors in a feature space. Symbolic concept representations, on the other hand, are amodal and language specific, and exist in the form of word / knowledge-graph embeddings in a concept / knowledge space. The human conceptual system comprises both embodied representations and symbolic representations, which typically interact to drive conceptual processing. As such, we further advocate the use of dual embodied-symbolic concept representations for deep learning. To demonstrate their usage and value, we discuss two important use cases: embodied-symbolic knowledge distillation for few-shot class incremental learning, and embodied-symbolic fused representation for image-text matching. Dual embodied-symbolic concept representations are the foundation for deep learning and symbolic AI integration. We discuss two important examples of such integration: scene graph generation with knowledge graph bridging, and multimodal knowledge graphs.

## 1 Introduction

Motivated by recent findings from *cognitive neural science*, we advocate the use of a *dual-level model for concept representations* [1-2]: the *embodied level* consists of *concept-oriented feature representations*, and the *symbolic level* consists of concepts in the form of *concept graphs*. For concrete concepts, the two levels are associated / connected. The embodied level corresponds to *sensorimotor-derived* knowledge representations of the dual-coding framework [4]; the symbolic level corresponds to *language-derived* knowledge representations of that framework.

*Embodied concept representations* are modality specific and exist in the form of *feature vectors* in a feature space. *Symbolic concept representations*, on the other hand, are amodal and language specific, and exist in the form of *word / knowledge-graph embeddings* in a concept / knowledge space. Traditionally, they are learned separately and used independently. For example, deep learning for *computer vision* learns embodied concept representations in the form of *visual feature vectors* for image classification, whereas deep learning for *natural language processing* learns symbolic concept representations in the form of *semantic word vectors* for sentiment analysis.

The *human conceptual system* comprises both simulated information of sensorimotor experience (i.e., *embodied representations*) and linguistic distributional information of how words are used in language (i.e., *symbolic representations*) [5-7]. Further, the *linguistic shortcut hypothesis* predicts that people will use computationally cheaper linguistic distributional information where it is sufficient for the task in question. However, people will resort to sensorimotor simulation to provide a more detailed, precise conceptual representation when required. Therefore, *both* symbolic and embodied representations are inherent to the functioning of the human conceptual system. Further, they typically interact to drive conceptual processing.

As such, we further advocate the use of *dual embodied-symbolic concept representations for deep learning*. To demonstrate their usage and value, we discuss two important use cases: *embodied-symbolic knowledge distillation* for few-shot class incremental learning (CIL) and *embodied-symbolic fused representation* for image-text matching. The first use case [16] demonstrates that embodied-symbolic knowledge distillation mitigates both the *catastrophic forgetting problem* for CIL and the *overfitting problem* for few-shot learning. The second use case [17] demonstrates that embodied-symbolic fused representation closes the *semantic gap* between images and text leading to improved performance when matching an image with text.

*Dual embodied-symbolic concept representations* are the *foundation* for deep learning and *symbolic AI* integration. We discuss two important examples of such integration: *scene graph generation with knowledge graph bridging*, and *multimodal knowledge graphs*. The first example [19] presents a unified formulation of scene graphs (SGs) and commonsense knowledge graphs (CSKGs), where a SG is seen as an *image-conditioned embodiment* of a CSKG. The second example [21] *grounds symbols* in knowledge graphs (KGs) to corresponding image, text, sound and video data and *maps symbols* to their corresponding referents with meanings in the physical world, which is a key step towards the realization of human-level AI.

## 2 Human Conceptual System: Symbolic and Embodied

The *human conceptual system* comprises both simulated information of sensorimotor experience (i.e., *embodied representations*) and linguistic distributional information of how words are used in language (i.e., *symbolic representations*) [5]. Further, the *linguistic shortcut hypothesis* predicts that people will use computationally cheaper linguistic distributional information where it is sufficient for the task in question. However, people will resort to sensorimotor simulation to provide a more detailed, precise conceptual representation when required. Therefore, *both* symbolic and embodied representations are inherent to the functioning of the human conceptual system, and they typically *interact* to drive conceptual processing.

In the following we discuss three recent findings in cognitive science that demonstrate the crucial role of *dual embodied-symbolic concept representations* in human cognition.

## 2.1 Category Production

A common way of testing how *concepts* are structured and accessed from long-term memory is with a *category production* task, whereby a participant is presented with a category name such as *'animal'*, and asked to name concepts belonging to that category. In a category production study [5], measures of *sensorimotor similarity* between a category and member concept (based on an 11-dimensional representation of sensorimotor strength) and *linguistic proximity* between the category name and member-concept name (based on word co-occurrence derived from a large corpus) are used to test category production performance. Both measures predict the order and frequency of category production, but *linguistic proximity* has an effect above and beyond sensorimotor similarity.

Sensorimotor and linguistic distributional information, therefore, are found to offer an explanation to the mechanisms driving responses in category production tasks [5]. In terms of *linguistic distributional information* (i.e., *symbolic information*), it is evident that the shared *linguistic contexts* between member-concept names and category name (e.g., between *cat / dog* and *'animal'*) in corpus-derived linguistic space is an effective predictor of category membership. In terms of *sensorimotor information* (i.e., *embodied information*), as suggested by many theories of conceptual structure, categorical distinctions emerge from common *features* (e.g., *fur*, *four-legged*) of member concepts that we perceive when interacting with the world. Furthermore, sensorimotor and linguistic distributional information typically *interact* to drive conceptual processing, and the *linguistic shortcut hypothesis* is the process by which people arrive at the most-frequently-named and first-named member concepts of a category.

## 2.2 Grounding of Concrete and Abstract Concepts

*Concrete concepts* (e.g., *animal*) have perceivable referents. In general, their semantic content can be clearly characterized, and their conceptual taxonomies can be unequivocally defined. Evidence obtained with concrete concepts [6] suggests interplay between modality-specific, multimodal and amodal semantic hub regions. Modality-specific and multimodal regions represent *conceptual features*, whereas amodal semantic hubs code *conceptual information* in an overarching supramodal fashion. The available evidence is most consistent with *hybrid models of conceptual representations* combining modality-specific and multimodal circuits with amodal conceptual hubs, i.e., *combining embodied and symbolic representations*.

*Abstract concepts* (e.g., *justice*) have referents which cannot be directly perceived, like mental or emotional states, abstract ideas, social constellations and scientific theories. Their semantic content is highly variable across individuals and contexts. As such, abstract concepts are a particular challenge for *embodied / grounded cognition theories* [6] because at the first glance, it is hard to imagine how abstract concepts, without a referent which can be perceived or acted upon, could be grounded in the sensorimotor brain systems. Embodied / grounded cognition theories, however, have been *refined* [6] in order to account for the representation of abstract concepts. Besides *sensorimotor information* due to metaphoric mapping or due to relations to classes of situations, the relevance of *emotional, introspective, social and linguistic information* has been stressed.

As is the case with concrete concepts, the findings with regard to *abstract concepts* [6] can be accommodated best by *hybrid models of conceptual representations* assuming an interaction between modality-specific, multimodal and amodal hub areas. *Modality-specific systems* include the sensorimotor systems, but also the modal systems involved in the processing of emotions, introspections, mentalizing, social constellations, and language. The latter modal systems are probably more important for abstract than for concrete concepts.

## 2.3 Meaning in Language and Mind

There are different theories on how much linguistic and sensorimotor representations contribute to meaning [7]. In *embodied theories*, meaning is based on the relation between words and *internal bodily states* corresponding to multiple modalities, such as vision, olfaction, and perhaps even internal affective states. By contrast, *symbolic theories* suggest that the meaning of word (e.g., rose) can be derived in a recursive fashion by considering its relation to the meaning of words in its *linguistic context* (e.g., red, flower). Current theories of semantics tend to posit that *both symbolic and embodied information contribute to meaning*.

A study [7] is made to evaluate how well different kinds of models account for people's representations of both concrete and abstract concepts. The models include *unimodal linguistic models* as well as *multimodal models* which combine linguistic with perceptual or affective information. There are two types of linguistic models considered: the *distributional linguistic model* which derives word meaning from word co-occurrences derived from large language corpora, and the *word association model* which measures the meaning of a word as a distribution of respectively weighted associative links encoded in a large semantic network. The study demonstrates that both *visual and affective multimodal models* better capture behavior that reflects human representations, especially for basic-level concepts that belong to the same superordinate category. The conclusion is that *multimodal information* (i.e., *symbolic and embodied information*) is important for capturing both abstract and concrete concepts.

# 3 Symbolic Concept Representations

*Embodied concept representations* (i.e., modality-specific *concept-oriented feature representations*) have been discussed previously [1-3]. In particular, we recommend their learning using *exemplar-based contrastive self-supervised learning (CSSL)* [3] since it is concept (class) centric and it supports class incremental learning.

In the following we discuss *symbolic concept representations* (i.e., amodal, language-specific *concept graphs*) since they are an integral part of *dual embodied-symbolic concept representations*, but we have not discussed them previously. They come in two forms: *word embedding* for representing *concepts* (namely, word semantics in natural language corpora or texts), and *knowledge graph embedding* for representing *concept graphs (a.k.a. knowledge graphs),* i.e., conceptual knowledge (e.g. commonsense knowledge).

Note that we will not discuss either word embedding learning or knowledge-graph embedding learning since they are of secondly importance to the focus of this paper. They are discussed in [9-10] and [12-14] respectively.

## 3.1 Word Embedding

*Word embeddings* [8-9] are dense representations of words in semantic vector spaces generated from *language corpora or texts*, in which semantically similar words have similar embedding vectors. Word embeddings have played an important role for tasks of *natural language processing* including complex and pertinent ones such as information retrieval and sentiment analysis. They are efficient to learn, highly scalable for large corpora (thousands of millions of words).

Two most widely used word embedding models [8-10] are Word2Vec and GloVe. *Word2Vec* employs two model architectures: Continuous Bag-of-Words (CBOW) model which aims to predict the occurrence of a word given other words that constitute its context, and Skip Gram (SG) model which deals with predicting a context given the word. Word2Vec considers only *local word co-occurrences. GloVe* is based on a model that reduces the dimensionality of a global co-occurrence matrix of the word-word type in a corpus, with the statistics of the entire corpus captured directly by the model. GloVe focuses on *global word co-occurrences*.

For *deep-learning based natural language processing* [10] word embedding is the basic building block that maps words in the input sentences into continuous space vectors, and usually used (pretrained) in the first layer of a neural network. Based on the word embedding, complex networks such as *recurrent neural networks (RNNs)* can be used for feature extraction and build, for example, *context-aware word embedding* or *phrase / sentence embedding*.

Note that the methods that generate word embedding based purely on information in a language corpus or text fail to take advantage of the semantic relational structure that exits between words in concurrent contexts. To overcome this limitation, the corpus or text is enhanced with extra morphological, syntactic, semantic and domain knowledge from knowledge sources (e.g., Wikipedia, Wordnet) to generate *knowledge-aware word embedding* [11].

## 3.2 Knowledge Graph Embedding

The *knowledge graph (KG)* [12-15] (a.k.a. *concept graph*) is a representation of conceptual knowledge (specifically, structured relational information) in the form of *concepts* and *relations* between them. We can view a KG as a set of statements (facts) having the form of *subject-predicate-object triples*, using the notation *(h, r, t) (head, relation, tail)* to identify a statement. We can also view a KG as a *directed labeled graph*, where *nodes* represent concepts and *edges* represent relations between concepts.

*Knowledge graph embedding (KGE)* [12-15] is a widely adopted approach to KG representation in which concepts and relations are embedded in *low-dimensional continuous vector spaces*. Most methods create *a vector for each concept and each relation*. These embeddings are generated in a way to capture latent properties of the semantics in the KG: similar concepts and similar relations will be represented with similar vectors. KGE offers precise, effective and structural representation of *symbolic conceptual information*.

Most of the KGE approaches rely mainly on the use of the *subject-predicate-object triples* present in the KG to generate the vector representations, *(h, r, t)*, for (head, relation, tail) [12-15]. These approaches can be broadly classified into two groups: translation-based models and semantic matching models. *Translation-based models* (e.g., TransE) are based on learning the translations from the head concept to the tail concept. They use *distance-based measures* to generate the similarity score for a pair of concepts and their relations. The training objective is to achieve, mathematically, **h** + **r** ~ **t**. *Semantic matching models* (e.g., RESCAL) use a multiplicative approach and represent the relations as *matrices / tensors* in the vector space. For example, RESCAL relies on a tensor factorization approach upon the 3-dimensional tensor generated by considering subject-predicate-object as the 3 dimensions of the tensor. Note that these models only consider each *individual fact*, while their intrinsic associations are neglected, which is not sufficient for capturing deeper semantics for better embedding. They, therefore, cannot meet the requirements of KGE.

New approaches leverage the *graph* nature of KG, and use *neural network (NN) based models* for various tasks [14-15]. When treated as a graph, KG can be seen as a heterogeneous graph, with the logical *relations* of more importance than the graph structure. NN-based models can consider the type of concept or relation, neighborhood / substructure information, path information, and temporal information. The use of *convolutional neural networks (CNNs)* or attention mechanisms also helps to generate better embeddings. Examples of *CNN-based models* include ConvE and ConvKB [14].

*Graph neural networks (GNNs)* are neural networks that can be directly applied to *graphs*, and provide an easy way to generate node-level, edge-level, and graph-level embeddings. They have a somewhat universal architecture in common, referred to as *Graph Convolutional Networks (GCNs)* which use deep, multi-layer processing known as *message passing*. Examples of *GCN-based models* include RGCN and SACN [14].

*GCNs* have a strong ability to mine the *underlying semantics* of KGs [14]. In general, GCN-based models can incorporate additional information, such as node types, relation types, node attributes and substructures, to generate better embeddings. For example, if the node information of a multi-hop domain can be aggregated, the accuracy of the model in specific tasks can be greatly improved.

## 4 Embodied-Symbolic Knowledge Distillation for Few-Shot CIL

In *CIL* [2], a model learns tasks continually, with each task containing a batch of new classes. In *few-shot CIL* [3], only the training set of the *first (base) task* may have large-scale training data for *base classes*, while other *subsequent (novel) task* just contains *few-shot* instances for *novel classes*. Few-shot CIL requires transferring knowledge (i.e., *knowledge distillation*) from old classes to new classes in solving the *catastrophic forgetting* problem, which is generic to CIL. The additional challenge of few-shot learning is the *overfitting* problem.

A knowledge distillation approach suitable for few-shot CIL is proposed in [16], which is based on the use of *dual embodied-symbolic concept (class) representations*. The *embodied (visual) representation* exists in the form of *visual vectors*. S*emantic word vectors* (Word2Vec and GloVe) are used as the *symbolic (semantic) representation* to facilitate *knowledge distillation*, which are generated from unsupervised learning on an unannotated *text corpus*. The semantically guided network does not add new parameters while adding new classes incrementally. The knowledge distillation process only includes *semantic word vectors* of novel classes in addition to the base classes, which are used to help the network remembering base class training, generalizing to novel classes, and generating well-separated embodied representation (*visual vectors*) of classes. Note that though the network may not have had the opportunity to see instances of novel classes, nevertheless, novel classes may very well share semantic properties with base classes it has seen. For example, if hyena is a novel class, many typical hyena attributes like 'face', 'body', etc., may have been seen by the network from base classes.

To mitigate the *overfitting* problem, *multiple visual embeddings* for classes are generated, where each is designed specifically for a group of classes. The *semantic word vectors* are used to separate classes into several groups. The number of groups / visual embeddings is defined by the *superclass (cluster)* knowledge obtained from the semantic word vector space. In the semantic space, there is a semantic word vector for each class. The set of superclasses is attained from the semantic word vector space representations of the *base classes*, and are then held fixed.

For the *first (base) task*, involving *base classes*, the steps for obtaining groups / visual embeddings are:

- Train the *network backbone* on the base classes, which is then kept frozen.
- Apply *k-means clustering*, where k = N, on *base semantic word vectors* and assign a *superclass (cluster) label* to each base class.
- Train *N embedding*s on the base task using superclass labels as group identity.

For other *subsequent (novel) task*, involving *novel classes*, the cluster centers (obtained in the base task) are used to assign *superclass labels* to novel classes. To assign a superclass label to novel classes, the minimum Euclidean distance between the semantic word vector of a novel class and cluster centers is used. Hence, given a novel class, there is a selection of *groups / visual embeddings* that each may be more or less suited.

An *attention module* is used to *merge* multiple visual embeddings of a class to generate its final visual embedding, i.e., *visual vector*. A *mapping module* is then used to project the visual vector from the visual space into the semantic space to *align* the visual vector with its associated *semantic word vector*. For the *first (base) task*, training involves attention loss and classification loss; for other *subsequent (novel) task*, training involves attention loss, distillation loss (i.e., alignment loss) and classification loss.

## 5 Embodied-Symbolic Fused Representation for Image-Text Matching

*Image and text matching* [17] is an important *vision-language cross-modality task* for many applications including image retrieval and caption. Before calculating the *similarity* between an image and text, a matching model needs to obtain a *rich representation* of the image (and text) first. Most of the current image-text matching models utilize pre-trained neural networks to extract feature embeddings as the representation of images. The *image embeddings*, however, fail to extract high-level semantic information. So the *semantic gap* between images and text leads to limited performance when matching an image with text.

Learning *semantic concepts* is useful to enhance image representation and can significantly improve the performance of both image-to-text and text-to-image retrieval. *Frequently co-occurred concepts* in the same image (scene), e.g. bedroom and bed, can provide commonsense knowledge to discover other semantic-related concepts. [17] uses a *Scene Concept Graph (SCG)* to support this by aggregating image scene graphs and extracting frequently co-occurred concepts as *commonsense knowledge*. Moreover, it proposes a novel model to incorporate this knowledge to improve image-text matching. Specifically, *semantic concepts* are *detected* from images and then *expanded* by the SCG to include *commonly-related concepts* (which may be occluded or long-tailed). Afterwards, it *fuses* their representations with the image embeddings, as *semantic-enhanced image embeddings*, to use (with text embeddings) for image-text matching.

The model uses *dual embodied-symbolic concept representations*. The *visual embodied representation* exists in the form of *image embeddings*. There are *two symbolic representations*. The *text symbolic representation* exists in the form of *(context-aware) word embeddings*. The *concept symbolic (semantic) representation* for images exists in the form of *concept embeddings*. Finally, there is an *embodied-symbolic (image-concept) fused representation* existing in the form of *concept-enhanced image embeddings*.

The *scene graph* [17] of an image is a graph consisting of concepts and relations between them. It can be represented as a set of triples of *<subject, relation, object>*. The *SCG* is constructed from scene graphs by aggregating *co-occurred <subject, object> pairs* from scene graphs of all images. To generate the concept embeddings for an image, first, a *concept detection* module (a multi-label image classification model) is used to extract *semantic concepts* from the image on a small concept vocabulary. With the SCG in hand, a *concept expansion* module is then used to expand the semantic concepts to include *commonly-related concepts*. Finally, a *concept prediction* module is used to predict *relevant concepts* from these and generate the *concept embeddings*. The concept embeddings of an image are fused with its *visual embedding*, by the *image-concept fusion* module, to generate a *concept-enhanced image representation*.

The *image-text matching* problem is formulated as a ranking model. Given the input image and the text, the output is the *similarity score* of matching their respective visual and language representations, with the visual representation being an *image-concept (embodied-symbolic) fused representation* and the language representation being *word embeddings*. To learn image and text matching as well as image-relevant semantic concepts jointly in an end-to-end fashion, the *loss function* consists of two parts: image-text matching loss and concept prediction loss.

# 6 Deep Learning and Symbolic AI Integration

*Symbolic AI* and *deep learning* both have strength and weakness, which tend to be each other's opposites. A significant challenge today [18] is to effect a *reconciliation*. Symbolic AI is based on manipulation of abstract compositional representations whose elements stand for *concepts and relations*. Therefore, to facilitate reconciliation, a key objective for deep learning is to develop architectures capable of discovering *concepts and relations* in raw data, and learning how to represent them.

An excellent example for doing this for image data has been discussed in 5 Embodied-Symbolic Fused Representation for Image-Text Matching. (Note that only *co-occurred concepts* are considered in [17]. However, their *relations* will also be considered in future work.) As discussed there, *dual embodied-symbolic concept representations* are the *foundation* for deep learning and symbolic AI integration.

In the following, we discuss two additional important examples of deep learning and symbolic AI integration.

## 6.1 Scene Graph Generation with Knowledge Graph Bridging

*Scene graphs (SGs)* are powerful representations that extract *semantic concepts and their relations* from *images,* which facilitate visual comprehension and reasoning. (For an example usage, see 5 Embodied-Symbolic Fused Representation for Image-Text Matching.) A SG can be represented as a set of triples of *<subject, relation, object>*. On the other hand, *commonsense knowledge graphs (CSKGs)* [20] are rich repositories that encode how the world is structured (i.e., *commonsense knowledge*), and how common *concepts* are *related* and interact.

*GB-Net (Graph Bridging Network)* [19] presents a unified formulation of these two constructs, where a SG is seen as an *image-conditioned embodiment* of a CSKG. Based on this perspective, the *SG generation* is formulated as the *inference of a bridge* between the SG and CSKG, where each *concept or relation* in the SG must be linked to its corresponding *concept or relation* in the CSKG.

Specifically, both SG and CSKG are defined as special types of knowledge graph (KG):

- A *KG* is a set of *nodes* of type *concept (C)* or *relation (R)*, and a set of directed, weighted *edges (E)* between the nodes.

- A *CSKG* is a type of KG with *commonsense concept (CC)* nodes and *commonsense relation (CR)* nodes. *Commonsense edges* are of four types: CC->CC, CC->CR, CR->CC, and CR->CR.
- A SG is a type of KG with *scene concept (SC)* nodes and *scene relation (SR)* nodes. *Scene edges* are of four types: SC->SR (subjectOf), SC->SR (objectOf), SR->SC (hasSubject), and SR->SC (hasObject).
- The SG and CSKG are connected through four types of *bridge edges*: SC->CC, SR->CR, CC->SC and CR->SR.

GB-Net uses *dual embodied-symbolic concept representations*. The *visual embodied representation* for images exists in the form of *image embeddings*. There are *two symbolic concept representations*. The *CSKG symbolic representation* exists in the form of *CSKG embedding* (of commonsense nodes and edges). The *SG symbolic (semantic) representation* for images exists in the form of *SG embedding* (of scene nodes and edges). The CSKG embedding and the SG embedding are *interconnected* through the bridge edges.

GB-Net *fuses* the SG embedding and CSKG embedding through a *dynamic message passing and bridging* algorithm using a *graph neural networks (GNN)*. The method iteratively propagates messages to update nodes, then compares nodes to update bridge edges, and repeats until the two graphs are carefully connected. This results in the *SG embedding* with *bridges to the CSKG embedding*.

## 6.2 Multimodal Knowledge Graphs

*Knowledge graphs (KGs)* have found great use in a wide range of applications including text understanding, recommendation system, natural language question answering, and *image understanding* (see 5 Embodied-Symbolic Fused Representation for Image-Text Matching and 6.1 Scene Graph Generation with Knowledge Graph Bridging). More and more KGs have been created, covering *common sense knowledge* [20] (see 6.1 Scene Graph Generation with Knowledge Graph Bridging), lexical knowledge, encyclopedia knowledge, taxonomic knowledge, and geographic knowledge.

Most of the existing *KGs* are represented with pure *symbols*, denoted in the form of text, without *grounding* to the physical world experience. However, *both symbolic and embodied representations* are inherent to the functioning of the *human conceptual system*, and they typically interact to drive conceptual processing (see 2 Human Conceptual System: Symbolic and Embodied). Therefore, it is necessary to *ground symbols* in KGs to corresponding *image, text, sound and video data* and *map* symbols to their corresponding referents with meanings in the physical world. That is, the multi-modalization of KGs [21] is an inevitable key step towards the realization of human-level AI, which results in *Multimodal Knowledge Graphs (MMKGs)* [12, 15, 21].

To support *symbol grounding* in MMKG [21], one can take *multimodal data* as particular *attribute values* of *concepts or relations*. This can be denoted in a *triple (s, r, d)*, where s denotes a concept or relation, d denotes one of its corresponding multimodal data, and the relation r is, e.g., "hasImage" when d is an image. Symbol grounding, therefore, can be divided into *concept grounding* and *relation grounding*.

As an example, *concept grounding* aims to find representative, discriminative and diverse *images* for *visual concepts*. A major challenge is to find *representative images* for a visual concept from a group of relevant images. The representativeness and discriminativeness of images can be scored in terms of results of *cluster-based methods*, such as K-means, based on *visual embeddings*. The *captions* of images can also be utilized to evaluate the representativeness and discriminativeness of images, at the *semantic level*, based on *text embeddings*.

IKRL and DKRL are two well-known examples of MMKG. *IKRL (Image-embodied Knowledge Representation Learning)* [12, 15] provides a method to integrate *images* inside the scoring function of the *KG embedding* model (TransE). Essentially, IKRL uses multiple images for each *concept* and use the AlexNet CNN to generate *embeddings for the images*. These embeddings are then selected and combined with the use of attention to be finally projected in the *KG embedding* space. *DKRL (Description-Embodied Knowledge Representation Learning)* [12, 15], on the other hand, includes the *description of concepts* in the representation. It uses a CNN to encode the concept description into a vector representation and uses this representation in the loss function. DKRL learns *two embeddings* for each concept, one that is *structure-based* (i.e., *KG*, like TransE) and one that is based on the *concept descriptions*. The two embeddings are *interconnected / integrated*.

The above discussion, though brief, shows that MMKG uses *dual embodied-symbolic concept representations*. The *KG symbolic representation* exists in the form of *KG embedding*. There are various *embodied representations*, depending on the *multimodal data* involved. For image data, the *visual embodied representation* exists in the form of *image embedding*. For text (descriptions), the *textual embodied representation* exists in the form of *text embedding*. The KG embedding and the modality-specific embedding(s) are *interconnected / integrated*.

# 7 Conclusion

Motivated by recent findings from cognitive neural science, we advocate the use of a dual-level model for concept representations: the embodied level consists of concept-oriented feature representations, and the symbolic level consists of concept graphs (a.k.a. knowledge graphs). The human conceptual system comprises both embodied representations and symbolic representations, which typically interact to drive conceptual processing. As such, we further advocate the use of dual embodied-symbolic concept representations for deep learning. That is, deep learning should learn from data not only modality-specific embodied representations such as image embeddings, text embeddings, etc., but also the corresponding amodal symbolic (semantic) representation as knowledge graph embeddings, with links to commonsense knowledge graphs. Dual embodied-symbolic concept representations are the foundation for deep learning and symbolic AI integration, which is an important direction for deep learning and AI since their integration reinforces each other's strength, compensates each other's weakness, and takes a major step toward human-level AI (e.g., grounding in experience / data, understanding, reasoning, explanation, etc.).

**Acknowledgement:** Thanks to my wife Hedy (郑期芳) for her support.